# Video Segment Copy Detection Using Memory Constrained Hierarchical Batch-Normalized LSTM Autoencoder

BTECH THESIS REPORT

*Submitted by*

**ARJUN KRISHNA**
15115030
**A S AKIL ARIF IBRAHIM**
15116001

*Under the guidance of:*

**Dr. VINOD PANKAJAKSHAN**

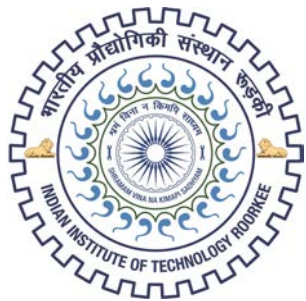

**DEPARTMENT OF ELECTRONICS AND COMMUNICATION ENGINEERING
INDIAN INSTITUTE OF TECHNOLOGY ROORKEE
ROORKEE − 247667 (INDIA)
May, 2019**

# ACKNOWLEDGEMENTS

We sincerely thank our B.Tech. Project advisor **Dr. Vinod Pankajakshan** for his constant guidance and support throughout the project.

This work was supported by **Signal Processing Lab at IIT roorkee** by providing us with the GPU workstation for training and experimenting with our deep learning model. We thank **Department of Electronics and Communication Engineering, IIT Roorkee** for providing the lab and other resources required for completing our work to fruition.

# ABSTRACT

In this report, we introduce a video hashing method for scalable video segment copy detection. The objective of video segment copy detection is to find the video (s) present in a large database, one of whose segments (cropped in time) is a (transformed) copy of the given query video. This transformation may be temporal (for example frame dropping, change in frame rate) or spatial (brightness and contrast change, addition of noise etc.) in nature although the primary focus of this report is detecting temporal attacks. The video hashing method proposed by us uses a deep learning neural network to learn variable length binary hash codes for the entire video considering both temporal and spatial features into account. This is in contrast to most existing video hashing methods, as they use conventional image hashing techniques to obtain hash codes for a video after extracting features for every frame or certain key frames, in which case the temporal information present in the video is not exploited. Our hashing method is specifically resilient to time cropping making it extremely useful in video segment copy detection. Experimental results obtained on the large augmented dataset consisting of around 25,000 videos with segment copies demonstrate the efficacy of our proposed video hashing method.

# TABLE OF CONTENTS

# LIST OF TABLES

# LIST OF FIGURES

# 1

## INTRODUCTION

## 1.1 Overview

In this chapter, we first discuss why the problem of video segment copy detection is important, and the *Motivation* of this report. Then we look at *Problem Definition*, and our *Contributions*. Finally, we describe the *Report Organization*.

## 1.2 Motivation

One of the most important requirements in digital media management is the detection of copies of media (audio, video and images). The applications of copy detection include usage tracking, copyright enforcement and information security. In this digital era, due to the abundance of video capturing equipments and video sharing platforms, the number of videos that are being transmitted online every day is increasing day by day. It is now easier than ever to copy a video and distribute it on the Internet. Hence video copy detection is an important and relevant problem today.

There are two techniques to detect copies of digital media, watermarking and content based copy detection. While watermarking has been researched more thoroughly, the field of content based copy detection is relatively new. In watermarking, some information which can identify the owner of the media is embedded into the media before its distribution. So every copy of this media contains the watermark, and this is then extracted to identify the owner of the media. The core concept behind content based copy detection is that *the media itself is the watermark*, i.e., the media (image, audio, video) possesses sufficient unique information which can be utilized for the task of copy detection.

The core idea that the content based copy detection schemes are based upon is the extraction of an unique signature from the original media and the measurement of the similarity between signatures extracted from a possible copy and the original media. The main merit of content based copy detection over watermarking is that even after the media has been distributed, the signature extraction can be done. For instance, a set of signatures can be created for a TV show, and then these signatures can be utilized to find all video clips from that TV show on the Internet. These clips can include screen recordings of the show, which will not contain the watermark that was originally embedded. Hence, watermarking approach cannot be used for this task. The content based copy detection can also be used in conjunction with watermarking. First, a content based copy detector is used to find possible copies of a media, which creates a suspect list that can be given to a creator or a distributor. Then the actual owner of the media can prove ownership by using watermark or other authentication techniques. There are several research efforts [1]-[6] that address content based copy detection.

Content based video copy detection could be used to prevent digital video piracy and copyright infringement. This issue has greatly worried content owners and producers throughout the years. The current laws attempt to tackle this issue by penalizing unlawful content service providers (like media sharing platforms), or people who download unlawful content. Similar measures are being adopted by governments all around the globe. For instance, in 2009, the State Administration of Radio, Film and Television (SARFT) of China created policies to take severe measures against unlicensed content distribution in the web [7].

One of the areas where watermarking cannot be used but content based copy detection could be used is to detect obscene, controversial and sensitive videos, which can then be removed by control agencies like the government. Swedish government has adopted laws against the Pirate Bay, a P2P file sharing website [8], [9], and UK government announced its plans to combat illegal P2P file sharing [10]. Since the scale of the Internet, as well as the dimensionality of the issue is huge, automated procedures have to be developed to find the aforementioned type of copied content. This report was conducted by keeping this key application in mind.

As mentioned above, the task of video copy detection is to accurately and efficiently detect the video from a large database that has the most relevant visual content, for a given query video. Videos are more complex than images in the sense that they also provide a high-level structured content across frames, in addition to the low-level visual content in every frame. Hence in comparison to image search, video search is a more challenging task. Furthermore, even a video of short duration, say 4 seconds with 25 frames per second contains $25 \times 4 = 100$ frames. This gives rise to a computationally expensive computation between frames, and hence is not practical for large-scale video databases. Therefore, the development of a framework for the task of scalable video copy detection which exploits discriminative information from videos remains an important problem.

Hashing is the conventional solution for scalable searching in a database. The main challenge on hashing methods is the generation of compact binary codes for representing the data points. The requirement of a good code is that similar data in the original space should be mapped to similar binary codes (similar in terms of hamming distance) and dissimilar data in original data to be well separated (in terms of hamming distance) in binary codes. To the best of our knowledge, there are no works which propose an efficient framework for video hashing for the purpose of video segment copy detection. Hence, we have explored hashing videos for the aforementioned task. Video hashing for content based retrieval can be classified into three categories. In the first category, a single representative feature vector is extracted for a video, and hashing is done. In the second category, each frame is treated as an image, and then image hashing is performed. The average of the resulting hamming distances between corresponding frames of two videos is taken. In the last category, several representative (key) frames are selected, and then image hashing is performed on these frames. Although these frameworks are straightforward, they are incapable of exploiting the inherent temporal information present in videos. Moreover, hashing the frames of a video individually is computationally expensive, since videos typically have a large number of frames. Therefore, it is essential to develop a framework for video hashing that utilizes the inherent temporal information present in a video to learn strong features, and subsequently also minimizes the computation of hamming distance.

To tackle the above challenges, we propose a video hashing method for scalable video segment copy detection, in which we make use of LSTM-based architecture [11]. Since Long Short Term Memory (LSTMs) have shown promising performance in several sequence learning tasks due to their capability of representing strong features for sequential input, we also use LSTM in our framework for video hashing.

## 1.3 Problem Definition

The aim of this report is to develop a scalable video hashing algorithm that can enable one to efficiently detect whether a given query video is a copy of a video segment present in the given large database. This report focuses on detecting video segment copies of varying durations.

## 1.4 Contributions

The summary of our contributions is as follows:

1. We propose a deep learning framework for video hashing which exploits the temporal and discriminative information present in videos, so that a video can be represented with representative and compact binary codes.

2. We conduct extensive scalable video copy detection experiments on the synthetic video segment copy dataset consisting of around 25,000 videos that we have created by augmenting Hollywood2 dataset to demonstrate the efficacy of our proposed video hashing method. Our method produces compact hashes for each video, and detects video segment copies of very short duration up to longer durations.

## 1.5 Report Organization

**Chapter 1** gives an introduction to the problem and the need for a video segment copy detection.

**Chapter 2** discusses the research work done in this problem area.

**Chapter 3** discusses the methodology proposed by us to tackle this problem.

**Chapter 4** provides the discussion of quantitative and qualitative results obtained.

**Chapter 5** concludes this report and provides the inferences drawn from this work and discusses the future scope of research in this area.

CHAPTER

## LITERATURE REVIEW

## 2.1 Overview

This chapter describes the various research works that are related to our work. Our work is related to *learning-based hashing*, *Image hashing*, *video hashing*, and *video hashing using deep learning* in terms of methodology and *video segment copy detection* in terms of application.

## 2.2 Learning-Based Hashing

Learning to hash [12] has been extensively used for the approximation of nearest neighbor search in the case of large-scale multimedia data, as it is computationally efficient and the retrieval quality is good. In learning-based hashing, a hash function $\boldsymbol{y} = \boldsymbol{h}(\boldsymbol{x}) \in \{0, 1\}^L$ (where $L$ is the length of the hash code) which maps an input data $\boldsymbol{x}$ to a compact code $\boldsymbol{y}$ is learned. This mapping leads to an efficient storage, and the search is computationally less expensive due to the fast bit XOR operations in Hamming space.

Many learning-based hashing methods like subspace models [13], [14], manifold models [15], [16], and kernel models [17], [18] have been studied extensively in the literature. These methods can be grouped into two categories: unsupervised [19]-[23] and supervised [24]-[30]. In the first category, label information is not required, and in the second category, class-wise label information is used. While most of these methods are nonlinear hashing techniques, only a handful of works utilized deep learning techniques. Furthermore, these methods are specifically developed for image retrieval, and cannot be applied directly for video hashing due to the inherent temporal information present in a video. A survey consisting of various hashing methods is given in [31]. In contrast, the focus of our work is on detection of copies of video segments using learning based hashing.

## 2.3  Image Hashing

There are two kinds of image hashing namely perceptual hashing (for example phash [32]) and cryptography hashing (for example Secure Hash Algorithm (SHA) [33]). The basic difference perceptual hash functions and cryptographic hashes is that perceptual hashes are similar if the features are analogous, whereas in cryptographic hashing, even a small change in the value of the input will create a substantial change in the value of the output. Perceptual hash functions are widely used in digital forensics and to find cases of online copyright infringement due to their ability to have a correlation between hashes, and so similar data can be found. Perceptual hashes are the type of image hashes relevant to this report. Perceptual hashes give similar hash codes (similar in terms of hamming distance) to two images that are perceived same as by the viewer. There are several methods for image hashes like the ones that use Discrete Cosine Transform (DCT) features [32], spectral hashing [34], etc. Detailed survey on recent perceptual image hashing techniques can be found in [35] and [36]. All the perceptual hash algorithms that we have studied have the same basic properties- the images can be resized, the aspect ratio can be changed, and there can even be small changes in the color of the image (brightness, contrast, etc.), and in spite of these changes, similar images will be matched. Image hashing techniques have been widely used for image forensics [37] but they cannot be readily extended to video hashing, as videos have temporal information also embedded in them.

## 2.4  Video Hashing

The existing video hashing algorithms can be grouped into three main categories:

1. **Frame-by-Frame Video Hashing** [38]-[40]: These techniques apply the image

hashing algorithm on all frames of the video, and the hash code of the video is the concatenation and/or combination of the resulting hashes of each frame. These techniques do not withstand simple temporal attacks like frame dropping.

2. **Key Frame Based Video Hashing** [41], [42]: These techniques identify few key frames for a video, and apply the image hashing algorithm on all the key frames of the video. The hash code of the video is the concatenation and/or combination of the resulting hashes of each key frame. These techniques are weak from a security perspective, i.e. they cannot detect the malicious modification of non-key frames. The performance of key frame based techniques for video copy detection tasks is heavily affected by the mismatch of key frame chosen for the original video from that of a copied query video. There are several techniques in key frame extraction that can be performed. Some of these are based on pixel-wise pixel color difference, likelihood ratio, histogram comparison, consecutive frame level difference etc.

3. **Spatio-Temporal Video Hashing** [43]-[45]: These techniques consider the input video as one single entity, and generate hash for a video by using its normalized version (via spatial re-sizing and temporal sub-sampling) which captures the temporal information along with the spatial information present in the video. This is essential as a video is much more than just a set of frames, and the task of video retrieval is more challenging than image hashing. The presence of temporal information in videos impedes the direct utilization of image-based hashing methods. Hence spatio-temporal video hashing outperforms frame by frame and key frame based video hashing.

   The attacks that video hashing algorithms have to be resilient to are of two types: spatial and temporal. Spatial attacks applied to a video are done in frame level, i.e all the frames are spatially altered the same way. Frame rotation, blurring, noise addition, brightness modification, contrast modification, modification of spatial resolution and addition of a logo are some example attacks in spatial domain. These kind of attacks are the similar to the ones encountered by image hashing techniques and are of less importance when it comes to sensitive video copy retrieval tasks. Temporal attacks like frame dropping, time cropping, changing the frame rates, etc. do not occur in case of image, and are of primary importance when it comes to sensitive video copy retrieval task.

## 2.5    Video Hashing Using Deep Learning

Recently,  there has been a dramatic improvement in the state-of-the-art in the areas of image classification, visual object detection and speech recognition, due to the application of deep learning. Motivated by this, works that employ deep learning for video hashing have started to come up. In [46], features are extracted from a variety of deep Convolutional Neural Networks (Deep ConvNets) (like RestNet [47] and VGG [48]). Cao et. al. [49] proposed HashNet, a novel deep learning architecture that learns hash codes by a continuation method. It learned the binary codes from imbalanced similarity labels but did not take into account the temporal information present in a video. Deep Pairwise-Supervised Hashing (DPSH) proposed in [50] considered the pairwise relationship and uses deep learning to extract the features and learn the hash codes simultaneously. Supervised Recurrent Hashing (SRH) [51] uses Long Short-Term Memory (LSTM) for modeling the structure of videos, and introduces a max-pooling technique to embed the frames into representations of fixed length, which are then utilized in the supervised hashing loss. Deep Video Hashing (DVH) [52] obtains the compact binary codes by using spatial-temporal information that is present after the stacked convolutional pooling layers for feature extraction. Self-Supervised Temporal Hashing (SSTH) [53] proposes a stacking strategy to integrate LSTM with hashing to generate hash codes that simultaneously capture the spatial and temporal information present in a video. In Self-Supervised Video Hashing (SSVH) [54], a hierarchical LSTM Autoencoder structure with fixed length (L) hashing is proposed, and uses nearest neighbour loss to account for hashing content-wise similar videos and reconstruction loss for optimized hash values.

## 2.6    Video Segment Copy Detection

Generally, there are two stages in a video copy detection technique: offline and online. During the offline stage, a unique descriptor (signature or fingerprint) is generated for every original video. The features are dependent on the content; either acoustic, visual, temporal or a combination of these form the descriptor. This type of techniques are also called Content Based Video Copy Detection (CBVCD) for this reason. Following the feature extraction stage, the descriptors are saved in a database. During the online step, similar features are extracted from the query video, and comparison of its descriptor with those of the original videos stored in the database is performed.

The intervention of deep learning for video copy detection [55]-[57] is very limited and most of the video copy detection in recent works [58], [59] deal with the scenario in which a reference video and a query video share very long copied segments (near duplicate detection). The prior arts that do try to deal with short video segments [60] use frame-by-frame or key frame based hashing techniques. But this leads to long length hash codes and therefore increased complexity in the online stage of copy detection. We therefore propose a deep learning based approach for video copy detection of short video segments based on spatio-temporal variable length video hashing.

# PROPOSED METHOD

## 3.1   Overview

The basic overview of the proposed method is given in Figure 3.1. Here, we train our neural net model and create the hash code database in the offline stage, and create hash code for the query video and compare it with the database to retrieve possible copies from within the database in the online stage as depicted in the diagram.

The rest of this section is divided as follows- We first introduce the *Notations Used and Problem Formulation*. Then we talk about the content based frame level *Feature Extraction*, followed by our novel framework, *Memory Constrained Hierarchical Batch Normalized Binary Auto-Encoder*. We then discuss about the *Loss Function* introduced to train our model and the *Extraction of Hash Codes* from our model. Finally, we discuss about the *Hash Code Comparison Strategy*.

## 3.2   Notations Used and Problem Formulation

A video $\boldsymbol{V}_j$ is represented as $\boldsymbol{V}_j = [\boldsymbol{F}_{j,1}, \boldsymbol{F}_{j,2}, \ldots, \boldsymbol{F}_{j,M_j}] \in \mathbf{R}^{M_j \times D}$, where $M_j$ is the number of frames in the video $\boldsymbol{V}_j$, and $D$ is the dimension of the features of every frame. $N$ denotes the number of videos in the train set. The features for every frame in a video are extracted in the feature extraction step, and $\boldsymbol{F}_{j,t}$ indicates the t-th frame features of the video $\boldsymbol{V}_j$.

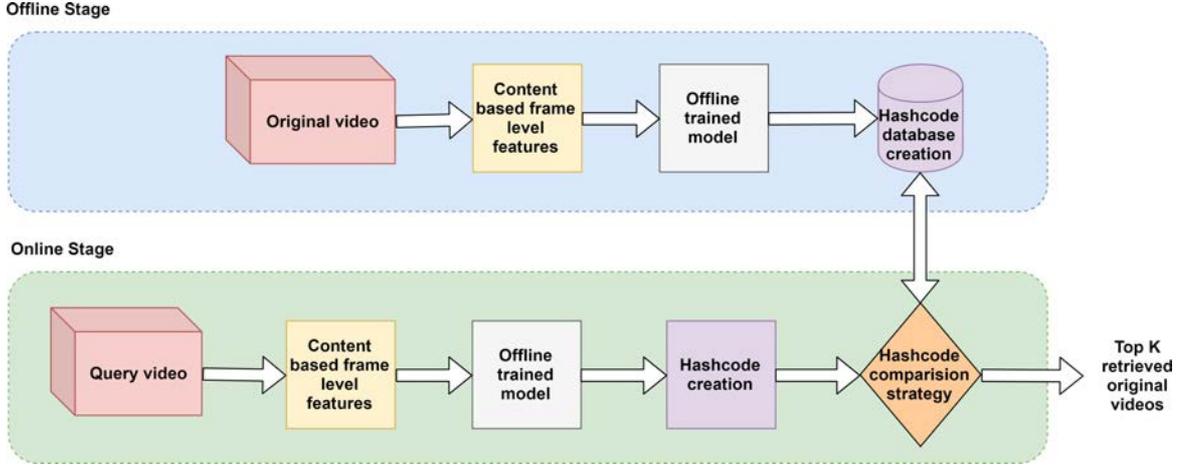

**Figure 3.1: Overview of proposed video segment copy detection**

$\boldsymbol{b}_j \in \{0, 1\}^{M_j \times L}$ denotes the binary outputs of the encoder for the video $\boldsymbol{V}_j$, where $L$ is the hidden state dimension of the last BNLSTM layer of the encoder.

We have to choose the optimum number of bits to represent a video. The trade off is that the more bits we use, the more information is conveyed but the search time will increase. We recognise *events* within a video, to solve the problem of optimum bit allocation. These events contain optimal amount of information for hash code creation. Note that this definition of event is different from event detection in literature (like [61]). We assign one fixed length hash code to each event. The aim of our video hashing method is to learn a binary code $\boldsymbol{c}_j \in \{0, 1\}^{E_j \times L}$ for each video $\boldsymbol{V}_j$, where $E_j$ is the number of events in the video $\boldsymbol{V}_j$. The hash code length of an event is given by $L$. $B$ denotes the mini-batch size.

## 3.3   Feature Extraction

We perform certain preprocessing and feature extraction on frame level to get frame level features. The basic overview of the same is given in Figure 3.2. We initially convert the frame rate of the videos to 25 frames per second, and obtain the frames. We reduce each frame to a $64 \times 64$ resolution gray scale image, and then compute $64 \times 64$ block DCT. The low frequency $32 \times 32$ components of DCT excluding the first DCT coefficient, are the features. We chose DCT features, as several perceptual image hash functions like pHash [33] use them. DCT have several desirable properties which make it a good candidate for perceptual image hashing. Low-frequency DCT coefficients of an image are invariant under several image manipulations [62]. This is because most of the useful information in the image tends to be concentrated in a few low-frequency components of the DCT. DCT coefficients are robust against attacks such as noising, compression, sharpening, and filtering. They are resilient to rotation up to a certain degree. However they are not resilient to extreme geometric attacks. Since the main focus of our work is on detecting video segment copies that have no or minimal geometric attacks, DCT coefficients are suitable features.

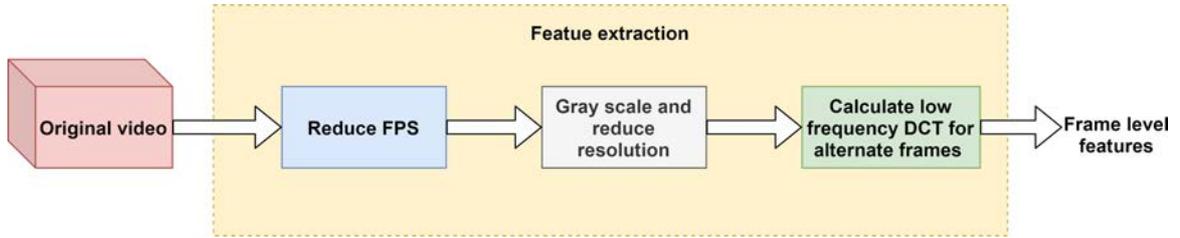

**Figure 3.2: Feature extraction**

To reduce the computational complexity, we take every alternate frame's feature as input, i.e. we are dropping every alternate frame. Each of these features is the input to the autoencoder model. We normalize the inputs with the mean and standard deviation of train dataset before feeding them to the autoencoder.

## 3.4 Memory Constrained Hierarchical Batch-Normalized LSTM Autoencoder

An Autoencoder is a neural network that is primarily used for dimensionality reduction. It is trained by backpropagation and provides an alternative to Principle Component Analysis (PCA) to perform dimensionality reduction by reducing the reconstruction error on the training set. It has proven to be an efficient architecture for tasks that require the input data to be projected to a latent space, for instance data dimensionality reduction [63], denoising [64] and hashing [11]. With regard to video hashing, auto encoders were used in [54]; however, there the purpose of video hashing was for not video segment copy detection. To the best of our knowledge, using autoencoders for the purpose of video segment copy detection have not been researched till now. Autoencoder in general has two modules- first the encoder which encodes the input data to a latent dimension (generally smaller than the input dimension) and a decoder that reconstructs the input

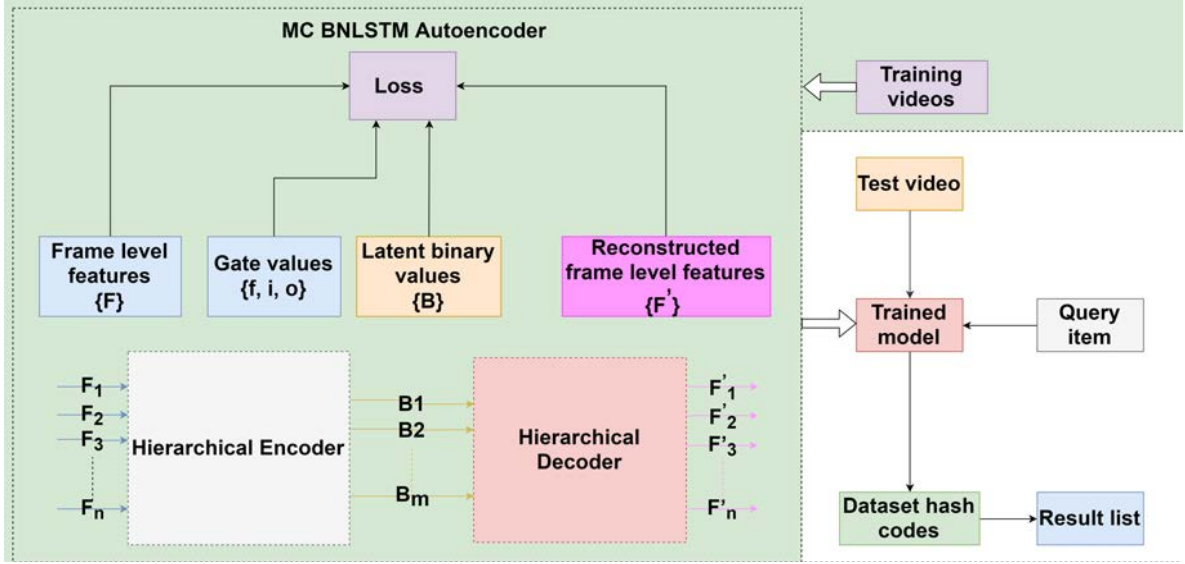

**Figure 3.3: Memory Constrained Hierarchical Batch-Normalized LSTM (MC BNLSTM) autoencoder (The left part with light green background shows the offline training stage and the right part shows the video segment copy retrieval process)**

data from the latent dimension. We have proposed Memory Constrained Hierarchical Batch-Normalized LSTM (MC BNLSTM) Autoencoder for the purpose of video segment copy detection. Figure 3.3 gives the overview of the same. The following part of this section explains the encoder and decoder module of our MC BNLSTM autoencoder.

## 3.4.1 Encoder

The architecture of the encoder is given in Figure 3.4. In the encoder, we have two BNL-STM (explained later on) layers followed by striding, then one BNLSTM layer followed by striding, and finally one BNLSTM layer.String pools information from neighbourhood frames and increases the perception of each LSTM cell in the deeper layers this enables the network to learn at varying granularities. In addition to this, striding reduces computational complexity [54]. The output of the last BNLSTM layer of the encoder is fed to a $\tanh^{-1}(\cdot)$ layer, to convert the range from $[-1, 1]$ to $(-\infty, \infty)$. This is done because the immediate binarization layer requires real valued input that ranges from $(-\infty, \infty)$. The output of this layer is fed to a binarization layer. The output of binarization layer will be used in the hash code extraction process. The length of the hash code of an event is the hidden dimension of the last BNLSTM layer of the encoder.

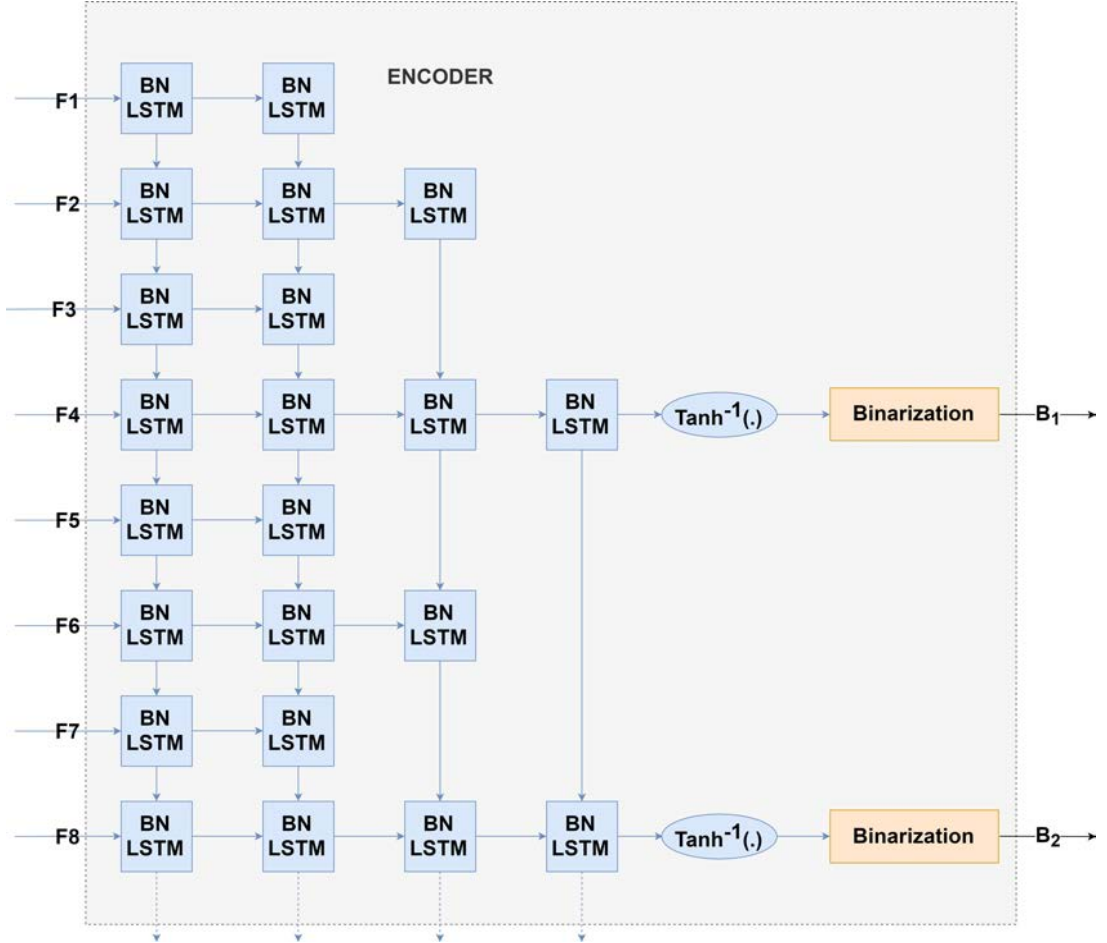

**Figure 3.4: MC BNLSTM Encoder architecture**

In a BNLSTM cell, batch normalization is done for each time step separately [66], so there is running mean and variance for all the time steps. Figure 3.5 compares cell of a time step for a normal LSTM cell to that of BNLSTM cell. If a test data is longer than the longest video in train dataset, then mean and variance for the time steps larger than the maximum time step in train dataset are taken to be the ones of the largest time step of train dataset. We batch normalize the weights and cell state. The batch normalizing transform is as follows [66]:

$$BN(\boldsymbol{h}; \gamma, \beta) = \beta + \gamma \odot \frac{\boldsymbol{h} - \mathrm{E}[\boldsymbol{h}]}{\sqrt{Var[\boldsymbol{h}] + \epsilon}} \qquad (3.1)$$

where $\boldsymbol{h} \in \mathbf{R}^d$ is the vector of (pre)activations to be normalized, $\gamma \in \mathbf{R}^d$, $\beta \in \mathbf{R}^d$ are model parameters (called affine parameters) that determine the mean and standard deviation of the normalized activation, and $\epsilon \in \mathbf{R}$ is a regularization hyper parameter. The division proceeds element wise.

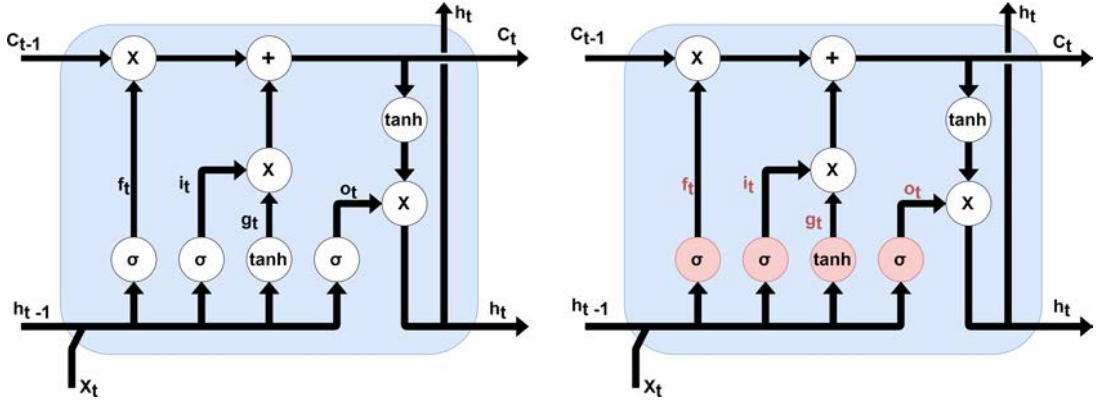

**Figure 3.5: LSTM cell on the left and BNLSTM cell on the right (the red colored portion of the BNLSTM cell are the quantities that are normalized for each time step)**

During training, the mean and standard deviation $\mathbf{E}[\boldsymbol{h}]$ and $Var[\boldsymbol{h}]$ are estimated by incorporating sample mean and sample variance of the current mini-batch in the running average and running mean. Hence we can differentiate through mean and standard deviation allowing for backpropagation through the statistics, this preserving the convergence of stochastic gradient descent (SGD). During inference of the model, the mean and variance are typically estimated based on the entire training set statistics, so as to produce a deterministic prediction. The batch-normalizing transform $BN(\cdot; \gamma, \beta)$ is introduced into the LSTM as follows [66]:

$$\begin{matrix} \boldsymbol{f}_t \\ \boldsymbol{i}_t \\ {}_t \\ \boldsymbol{g}_t \end{matrix} = BN(\boldsymbol{W}_h \boldsymbol{h}_{t-1}; \gamma_h, \beta_h) + BN(\boldsymbol{W}_x \boldsymbol{x}_t; \gamma_x, \beta_x) + \begin{matrix} \boldsymbol{b} \\ \boldsymbol{o} \end{matrix} \qquad (3.2)$$

$$\boldsymbol{c}_t = \sigma(\boldsymbol{f}_t)^- \boldsymbol{c}_{t-1} + \sigma(\boldsymbol{i}_t)^- tanh(\boldsymbol{g}_t) \qquad (3.3)$$

$$\boldsymbol{h}_t = \sigma(\boldsymbol{o}_t)^- \tanh(BN(\boldsymbol{c}_t; \gamma_c, \beta_c)) \qquad (3.4)$$

where $\boldsymbol{f}_t$, $\boldsymbol{i}_t$ and $\boldsymbol{o}_t$ denote the values of forget gate, input gate and output gate and at t-th time step respectively, $\boldsymbol{W}_h \in \mathbf{R}^{d_h \times 4d_h}$, $\boldsymbol{W}_x \in \mathbf{R}^{d_x \times 4d_h}$, $\boldsymbol{b} \in \mathbf{R}^{4d_h}$ and the initial states $\boldsymbol{h}_0 \in \mathbf{R}^{d_h}$, $\boldsymbol{c}_0 \in \mathbf{R}^{d_h}$ are model parameters. $\sigma$ is the logistic sigmoid function, and the $^-$ operator denotes the Hadamard product.

The recurrent term $\boldsymbol{W}_h \boldsymbol{h}_{t-1}$ and the input term $\boldsymbol{W}_x \boldsymbol{x}_t$ are normalized separately, since normalizing these terms individually gives the model better control over the relative contribution of the terms using the $\gamma_h$ and $\gamma_x$ parameters. We set $\beta_h = \beta_x = 0$ to avoid unnecessary redundancy, instead relying on the pre-existing parameter vector **b** to account for both biases. In order to leave the LSTM dynamics intact and preserve the gradient flow through $c_t$, batch normalization is not applied in the cell update. During training, the mean and variance across the mini-batch are estimated independently for each time step. During testing, the estimates for mean and variance to be used are obtained by averaging the mini-batch estimates over the entire training set.

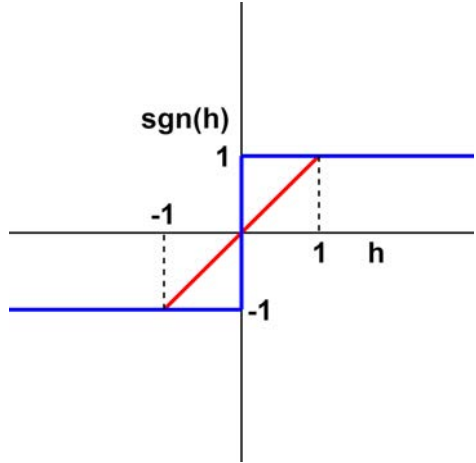

**Figure 3.6: Approximation for signum function (red graph) by sgn(h) (blue graph)**

In the binarization layer, we cannot use the signum function to binarize the inputs to this layer. This is due to the fact that the derivative of the signum function is zero everywhere except at 0, and hence it is impossible to apply exact backpropagation. Therefore, we use an approximation of the signum function (Figure 3.6) by estimating the derivative of the signum function[53]:-

$$sgn(h) = \begin{cases} -1, & h < -1 \\ h, & -1 \le h \le 1 \\ 1, & h > 1 \end{cases} \tag{3.5}$$

Therefore, $sgn'(h) = \mathbb{1}(|h| \leqslant 1)$, where $\mathbb{1}(.)$ is the indicator function.

### 3.4.2 Decoder

The architecture of decoder is depicted in Figure 3.7. The binary output of encoder is fed to the first layer of decoder. In the decoder, we have one BNLSTM layer followed by upsampling. The upsampling that we use here is average upsampling where we average the input of previous and the next time step to get the input to the current time step. After this layer, we have a BNLSTM layer followed by upsampling (upsampling is done the same way as mentioned before), and finally two BNLSTM layers. Then the output of this layer is fed to a $\tanh^{-1}(\cdot)$ layer. This is done because the output of the BNLSTM lies between $[-1, 1]$ whereas the reconstructed frame level features should lie between $(-\infty, \infty)$.

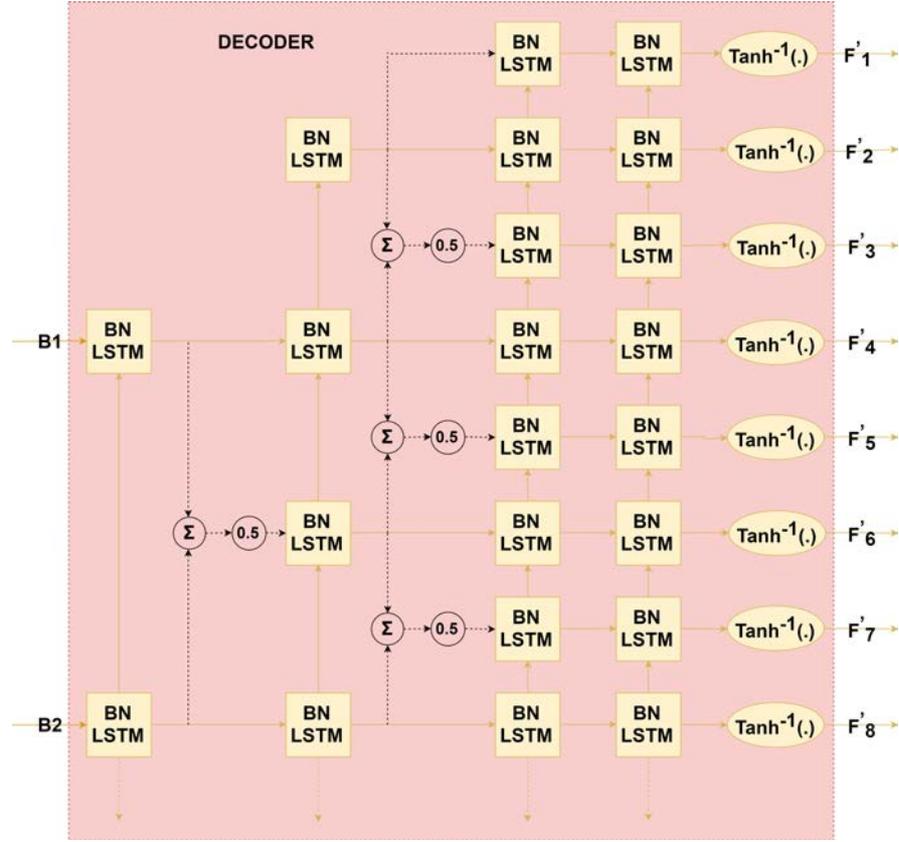

**Figure 3.7: MC BNLSTM Decoder architecture (The dotted arrows show the average upsampling performed in the decoder)**

## 3.5 Loss Function

This section formulates the loss function which is used to train our model. Let $\theta_e$ and $\theta_d$ denotes the parameters of encoder and parameters of decoder respectively. The binary outputs for the video $\boldsymbol{V}_j$ can be obtained by encoder as:

$$\boldsymbol{b}_j = Encoder(\theta_e, \boldsymbol{F}_j) \qquad (3.6)$$

Then $\boldsymbol{b}_j$ is used to reconstruct video frame features $\boldsymbol{V}_j^{\emptyset} = [\boldsymbol{F}_{j,1}^{\emptyset}, \boldsymbol{F}_{j,2}^{\emptyset}, \ldots, \boldsymbol{F}_{j,M_j}^{\emptyset}]$ in the decoder.

$$\boldsymbol{V}_j^{\emptyset}, \boldsymbol{f}_j, \boldsymbol{i}_j, \boldsymbol{o}_j = Decoder(\theta_d, \boldsymbol{b}_j) \qquad (3.7)$$

where $\boldsymbol{f}_j$, $\boldsymbol{i}_j$ and $\boldsymbol{o}_j$ are the values of forget gate, input gate and output gate of the last layer in encoder respectively. The total loss is weighted sum of the three loss functions- *Reconstruction loss*, *Memory Loss*, and *Diversity Loss*. These loss functions are explained below.

### 3.5.1   Reconstruction loss

This is the Mean Square Error (MSE) between the reconstructed output and input of autoencoder. This is the general loss used for training autoencoders for dimensionality reduction. Reconstruction for the video $\boldsymbol{V}_j$ is given by:

$$ReconLoss_j = \frac{1}{L \times M_j} \sum_{t=1}^{M_j} \left\| \boldsymbol{F}_{j,t} - \boldsymbol{F}_{j,t}^{\emptyset} \right\|^2 \qquad (3.8)$$

This should be minimized during training time. This is scaled by the hash size of an event $L$ and the number of frames $M_j$ in the video $\boldsymbol{V}_j$.

### 3.5.2   Memory loss

As discussed in Sec.3.2, a video is a sequence of events, where each event contains the optimal amount of information for creation of hash code of the video. The events must be independent of each other, so that the hash codes that they receive are also independent. So, if a copy contains only one of these events, the hash code assigned to this copy is similar to the hash code of the corresponding event. Hence at the start of an event, we want our network to forget the past inputs and consider only the current input. So the L2 norm of forget gate must be very low, and that of input gate must be high. L2 norm of output gate should also be low, since output gate depends on output of previous time step. During an event, the outputs of adjacent time steps are similar, so to stabilize the outputs of the time steps, L2 norm of forget gate must be maximized and that of input gate should be minimized, i.e. more weightage is given to past inputs compared to current input. L2 norm of output gate should also be maximized. Due the memory constraint (for forget ($\boldsymbol{f}_j$), input ($\boldsymbol{i}_j$) and output ($\boldsymbol{o}_j$) gates) that we bring about in our loss formulation, we have coined our architecture as Memory Constrained Hierarchical Batch-Normalized LSTM (MC BNLSTM) autoencoder.

Let $\boldsymbol{d}_{j,t}$ denote the hamming distance between $\boldsymbol{b}_{j,t+1}$ and $\boldsymbol{b}_{j,t}$. If $\boldsymbol{d}_{j,t}$ exceeds a predefined threshold $th$ for some $t^l$, then there is definitely an event transition at the time step $t^l$. So we have formulated our memory loss such that once the $\boldsymbol{d}_{j,t}$ exceeds a certain threshold the memory constraint discussed in the above paragraph while an event is transitioning has to be minimised and when $\boldsymbol{d}_{j,t}$ is less than a certain threshold the memory constraint discussed in the above paragraph within an event has to be minimized. The number of time steps where event transitions happen are small in comparison to the number of frames in the video. There is a multiplication factor of $\boldsymbol{d}_{j,t}$ present in the loss function to increase the weightage of the time steps where $\boldsymbol{d}_{j,t} \geq th$, and to decrease the weightage of other time steps. Memory loss for the video $\boldsymbol{V}_j$ is given by:

$$
\begin{aligned}
MemLoss_j = \ & \frac{1}{L^2 \times M_j} \sum_{t=1}^{M_j-1} \mathbb{1}(\boldsymbol{d}_{j,t} \geq th)(\|\boldsymbol{f}_{j,t}\|^2 + \|\boldsymbol{o}_{j,t}\|^2 + \|1 - \boldsymbol{i}_{j,t}\|^2)\hat{\boldsymbol{d}}_{j,t} \\
& + \frac{1}{L^2 \times M_j} \sum_{t=1}^{M_j-1} \mathbb{1}(\boldsymbol{d}_{j,t} < th)(\|1 - \boldsymbol{f}_{j,t}\|^2 + \|1 - \boldsymbol{o}_{j,t}\|^2 + \|\boldsymbol{i}_{j,t}\|^2)\boldsymbol{d}_{j,t}
\end{aligned}
\tag{3.9}
$$

This loss is scaled to lie between 0 and 1.

### 3.5.3  Diversity loss

We want the hash codes of different videos to be different. So, we want to maximize the hamming distance between them. This loss gives the measure of similarity between binary outputs of the encoder for different videos in a mini-batch. This loss must be minimized. This loss is scaled to lie between 0 and 1. Diversity loss is given by:

$$
DivLoss = \frac{1}{\sum_{j=1}^{B-1}\sum_{k=j+1}^{B} M_j \times M_k} \sum_{j=1}^{B-1}\sum_{k=j+1}^{B}\sum_{t_j=1}^{M_j}\sum_{t_k=1}^{M_k} Hamming(\boldsymbol{b}_{j,t_j}, \boldsymbol{b}_{k,t_k})
\tag{3.10}
$$

where $Hamming(\boldsymbol{b}_{j,t_j}, \boldsymbol{b}_{k,t_k})$ denotes the hamming distance between $\boldsymbol{b}_{j,t_j}$ and $\boldsymbol{b}_{k,t_k}$.

### 3.5.4 Total Loss

The total loss is given by:

$$TotalLoss = \frac{1}{N} \sum_{j=1}^{N} (ReconLoss_j + MemLoss_j) + DivLoss \qquad (3.11)$$

We can update the parameters $\theta_e$ and $\theta_d$ by making use of back propagation to optimize our model.

## 3.6   Extraction of Hash Codes

Since videos of different duration contain varying levels of information, it is necessary to make use of variable length hashing for the whole video. However the concept of events are granular in nature and is independent of event duration. Therefore we use variable length hashing, where one fixed length hash code is assigned to each event. The binary outputs of the encoder are used to extract the hash code of the given video. We find the time steps where the hamming distance between outputs of adjacent time steps exceeds a threshold, or exceeds the local average hamming distance between adjacent time steps. These time steps denote the ends of an event. The hash code assigned to an event is the majority pooling of output of the end of the event with outputs of few other previous time steps. The hash code of a video is concatenation of hash codes of its events.

## 3.7   Hash Code Comparison Strategy

This section talks about finding minimum hamming distance between hash code of the given video with hash code of a video in the database. Since the main focus of our project is on compact representation of hash codes, optimization of hash code comparison strategy is beyond the scope of this report. For our case, we have used a simple comparison strategy wherein the hash code of each event of the given query video $V_q$ is compared with hash code of each event of the video $V_d$ in the database, and the minimum hamming distance between each event of $V_q$ and $V_d$ is computed by maximal substring matching. The time complexity of the comparison strategy is $O(L_d \times L_q)$, where $L_d$ and $L_q$ are the number of events in the videos $V_d$ and $V_q$ respectively. The minimum hamming distance between the videos $V_d$ and $V_q$ is the average of minimum hamming distances between each event of $V_q$ and $V_d$. This is done for all the videos in the database.

The $k$ videos in the database with the least minimum hamming distances are retrieved. The given query video might be a part of these $k$ videos.

# RESULTS AND DISCUSSION

## 4.1   Experimental Setup

All our experiments were conducted with PyTorch [67] on four NVIDIA 1080 Ti GPUs. The multi-GPU training is done in such a way that there is one copy of the network in each GPU, with separate batch normalization affine learnable parameters $\gamma_h$, $\beta_h$, $\gamma_x$, $\beta_x$, $\gamma_c$ and $\beta_c$. The mini-batch is scattered in these 4 GPUs, and the forward pass is carried out. Then the loss is calculated in each GPU and backpropagated independently. The gradients hence obtained in each GPU are averaged together, and parameter updation is done completing one iteration of training in an epoch.

## 4.2   Dataset

We chose Hollywood2 dataset [68] for evaluating the performance of our proposed method. This dataset provide videos for the purpose of human action recognition in challenging and realistic settings. This dataset contains 3669 video clips of approximately 20.1 hours of video in total, divided into 10 class of scenes and 12 class of human action. The videos in this dataset are from video clips taken from 69 movies. This dataset is split into two parts with scene samples and actions. Since our problem just requires video clips, we used only the scene samples. The average duration of a video clip is around 35 seconds. The dataset was split into train set of 1000 videos (so $N = 1000$) and test set of

100 videos.

## 4.3   Implementation Details

First, feature extraction is done to get the frame level features, whose dimension $D$ is 1024. Then, every alternate frame's features are taken as input to our network. The dimension of hidden state of BNLSTM layers were chosen to be 256, 256, and 64 for the first 3 layers and dimension of the hidden state of last BNLSTM layer is same as the hash code length for an event (denoted by $L$). This hash code length $L$ was chosen to be 64. The decoder is the mirror image of the encoder, and therefore has the layer dimensions in the reverse order of that of encoder. As discussed in Sec.3.4.1, striding is performed at two places in the encoder. The stride at both of these places was chosen to be 2, and the same is done in the case of decoder also. During training, we chose Adam optimizer [69] with $\beta_1 = 0.9$, $\beta_2 = 0.999$ and $\epsilon = 1e - 8$, as they are mentioned as good default settings in [69]. The mini-batch size $B$ was taken to be 32.

## 4.4   Evaluation Metrics

We use the top-$k$ accuracy [70] to evaluate our model. The key problem that we are trying to address using the concept of events is to reduce the number of misdetections due to frame mismatch. Therefore to compare the performance of our model to a methodology where we select hashes in some fixed interval duration, we define three types of hash codes:

1. **Events:** These are the hash codes obtained by concatenating the hash codes for the time steps where events are detected by our model.

2. **Sample:** These are the hash codes obtained by generating one hash code for every $T_s$ seconds, where $T_s$ is the sampling duration. We take $T_s$ to be 4 seconds in our experiments.

3. **Sample and Events:** We first find the time steps where there is an event end. Then, if time steps of two adjacent event ends, say $T_i$ and $T_j$, differ by a value greater than or equal to $2T_s$ seconds, we append the time step $T_i + T_s$ with the time step of event ends, and repeat this procedure again.

We compare the results obtained from these three types of hash codes. The following is done for each type of hash code:-

1. After the training is complete, we extract the hash codes for all videos of train set and store it in a lookup table. For each video in the train set, we extract copies of various start and end time steps, and of various time durations. Then we obtain the hash codes of all these copies and store it in a lookup table. The next point describes how these copies are created in detail.

2. *Creation of copies*:- We take minimum time duration of a copy to be 4 seconds. The duration of copies are multiples of this minimum duration. The entire video is broken into $T_{f\ v}/T_c$ number of copies, where $T_{f\ v}$ denotes the duration of the full video and $T_c$ denotes the duration of the copy. Hash codes of each of these copies are obtained. Then, the start time step of the full video is changed to 2 seconds, and we define this to be the *minimum slide*. The hash codes of these copies are found. We repeat the above procedure by changing the start time step of the full video to multiples of minimum slide. The synthetic dataset created in this way contains around 25,000 videos which leads to a very exhaustive list of video segment copies of varying duration and varying slide. The number of comparisons hence done is of the order of $25,000 \times 1000$, which is 25 million.

3. Then, minimum hamming distance between every copy and every video in the train set is found. Then for every copy, we retrieve $k$ videos from the train set with the least minimum hamming distance. If the original video is present in these $k$ videos, then the copy is successfully detected, else it is not detected. We find the top-$k$ accuracy for the values of $k$ from 1 to 10. The resulting curves for the train set are given in Figure 4.1.

For the test set, the above mentioned procedure is repeated except that now the minimum hamming distance is found between a partial copy and all videos present in both train and test set. So the retrieved $k$ videos are now from both the train and the test set. The resulting curves for the test set are given in Figure 4.3.

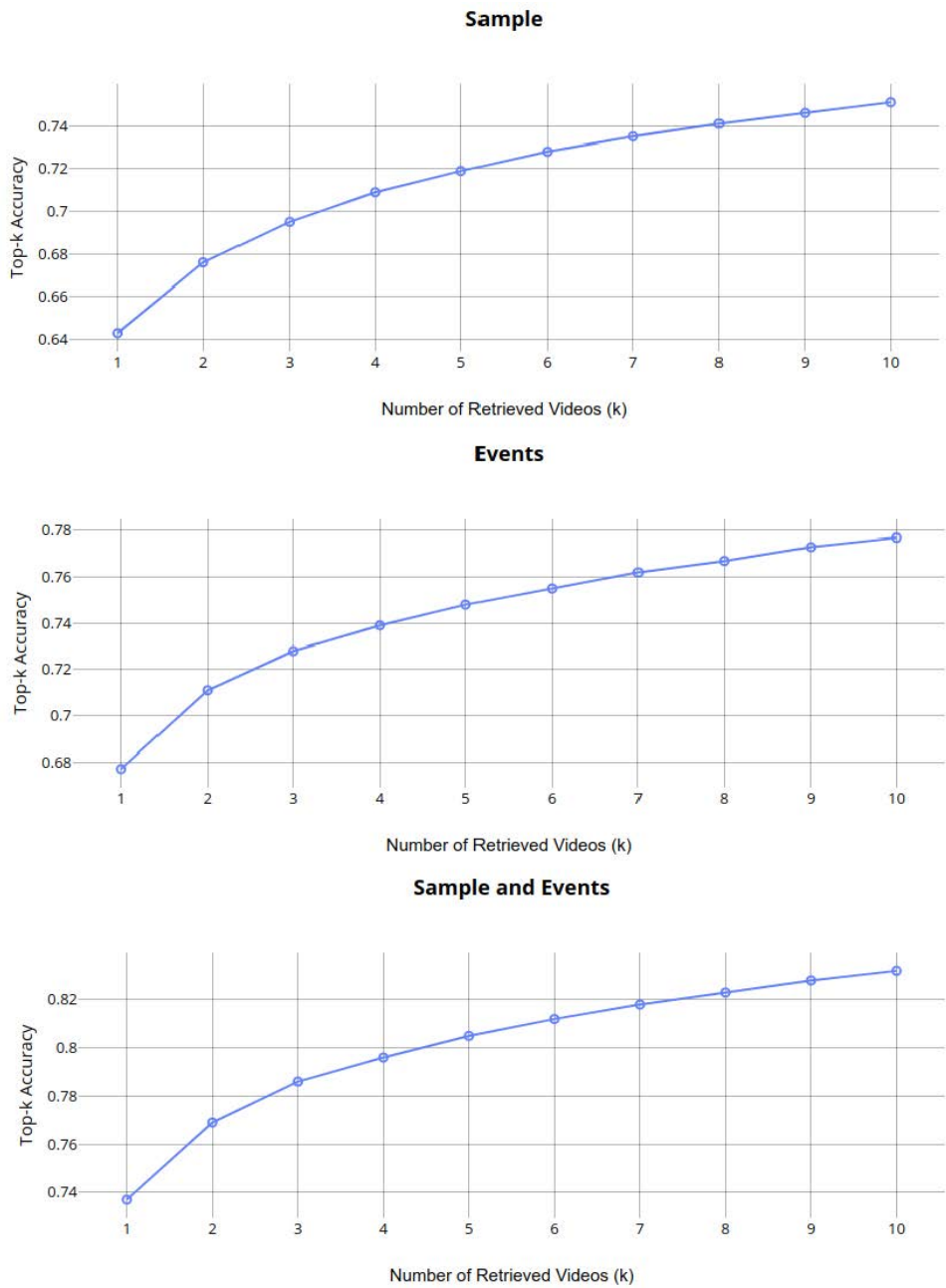

**Figure 4.1: Plot of Top-$k$ Accuracy for Train Set**

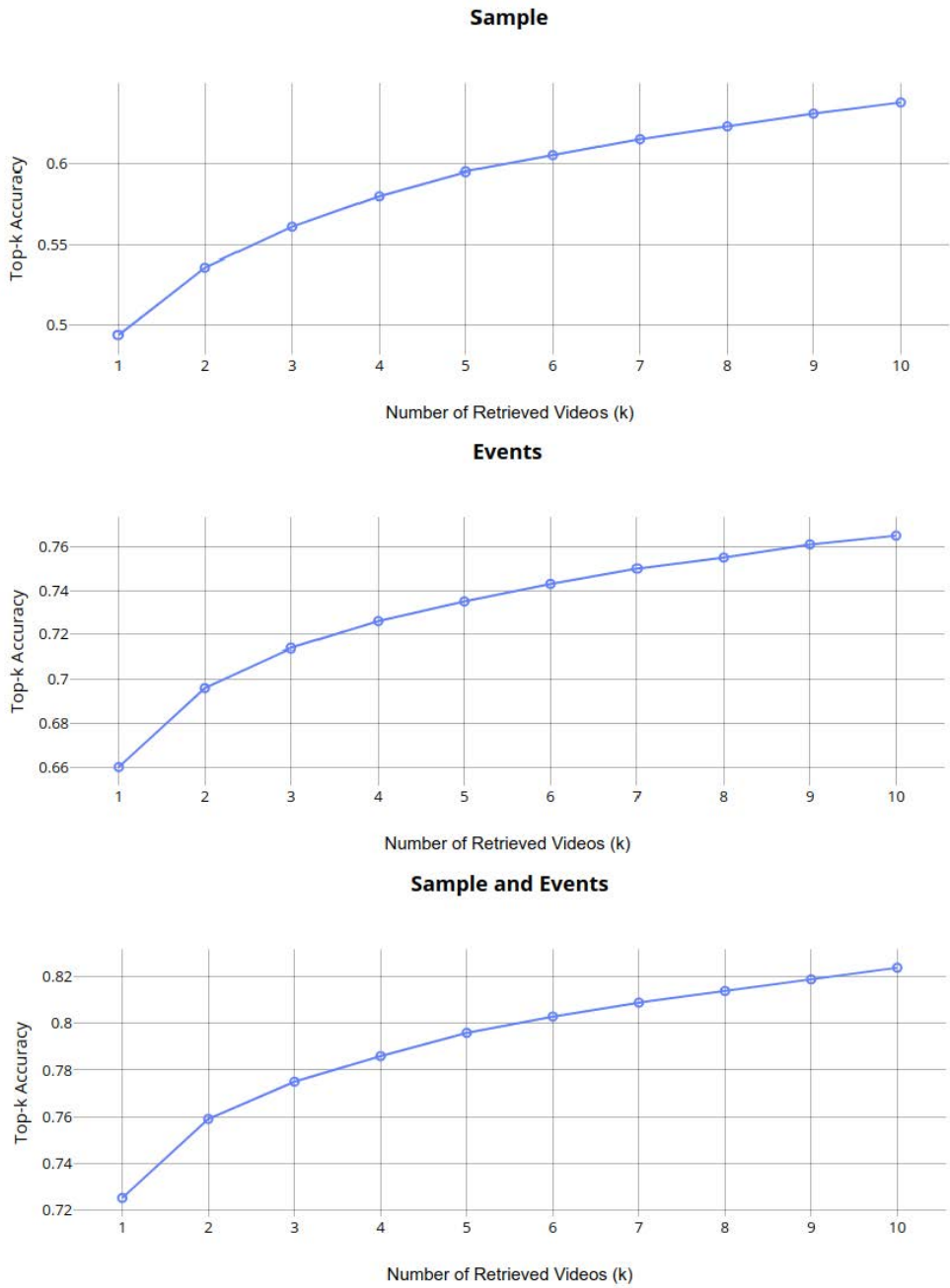

**Figure 4.2: Plot of Top-*k* Accuracy for Train Set for slide values that are a multiple of 2 seconds but not 4 seconds**

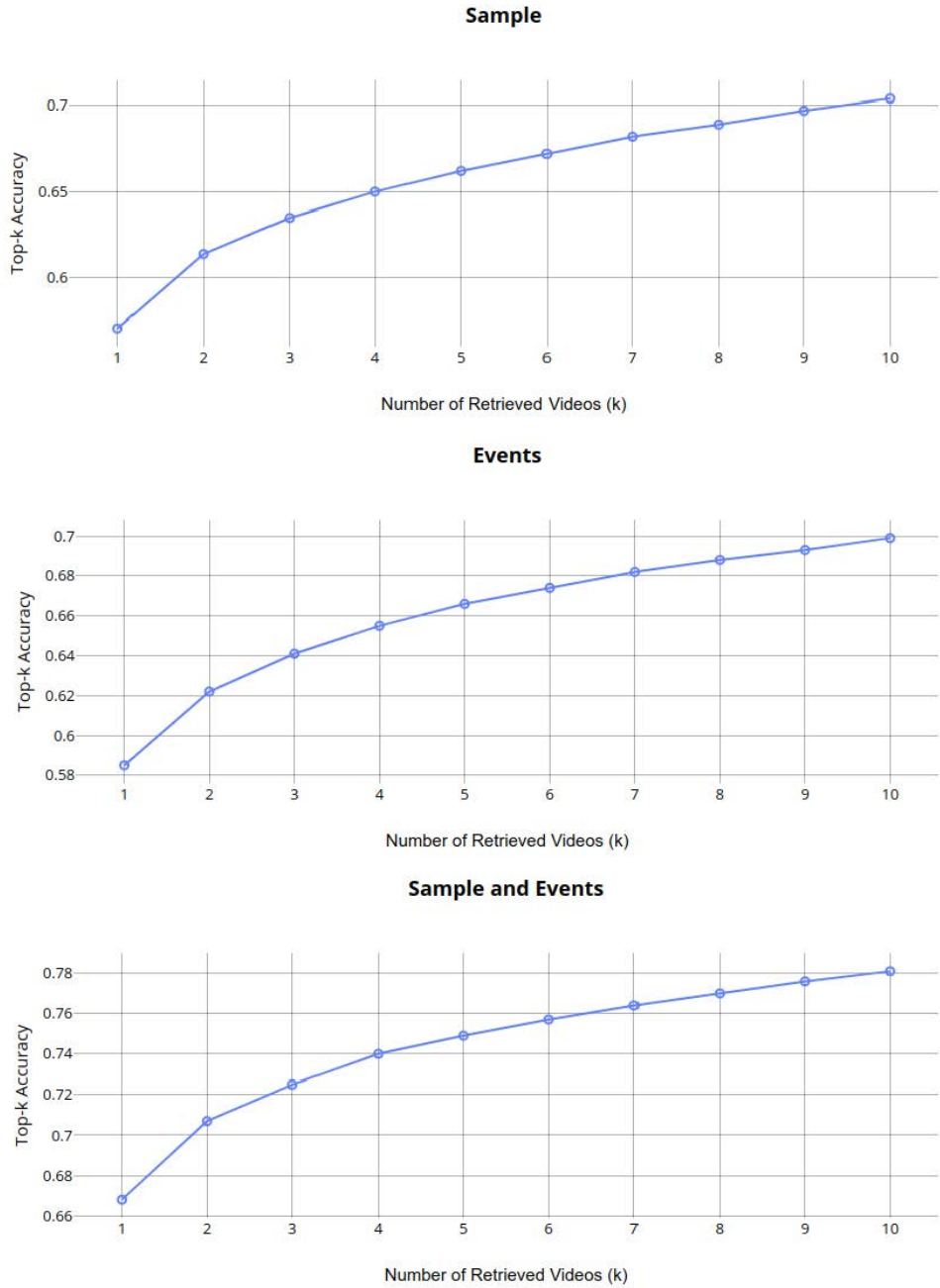

**Figure 4.3: Plot of Top-*k* Accuracy for Test Set**

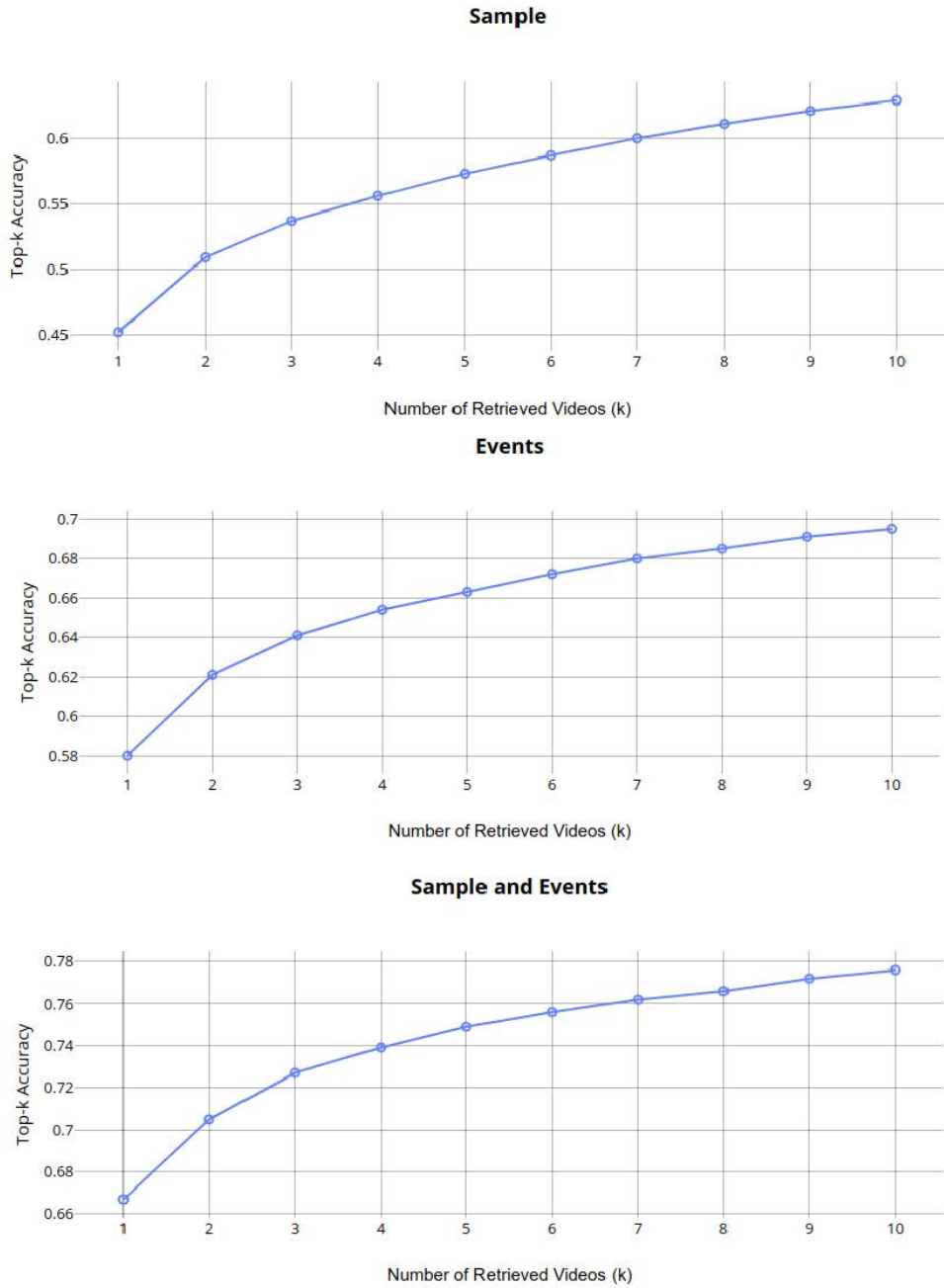

**Figure 4.4: Plot of Top-$k$ Accuracy for Test Set for slide values that are a multiple of 2 seconds but not 4 seconds**

**Table 4.1: Top-*k* Accuracy for Train Set**

| k | Sample | Events | Sample and Events |
|---|--------|--------|-------------------|
| 1 | 0.643 | 0.677 | 0.737 |
| 2 | 0.676 | 0.711 | 0.769 |
| 3 | 0.695 | 0.728 | 0.786 |
| 4 | 0.709 | 0.739 | 0.796 |
| 5 | 0.719 | 0.748 | 0.805 |
| 6 | 0.728 | 0.755 | 0.812 |
| 7 | 0.735 | 0.762 | 0.818 |
| 8 | 0.741 | 0.767 | 0.823 |
| 9 | 0.746 | 0.773 | 0.828 |
| 10 | 0.751 | 0.777 | 0.832 |

**Table 4.2: Top-*k* Accuracy for Train Set for slide values that are a multiple of 2 seconds but not 4 seconds**

| k | Sample | Events | Sample and Events |
|---|--------|--------|-------------------|
| 1 | 0.494 | 0.66 | 0.725 |
| 2 | 0.536 | 0.696 | 0.759 |
| 3 | 0.561 | 0.714 | 0.775 |
| 4 | 0.58 | 0.726 | 0.786 |
| 5 | 0.595 | 0.735 | 0.796 |
| 6 | 0.605 | 0.743 | 0.803 |
| 7 | 0.615 | 0.75 | 0.809 |
| 8 | 0.623 | 0.755 | 0.814 |
| 9 | 0.631 | 0.761 | 0.819 |
| 10 | 0.638 | 0.765 | 0.824 |

**Table 4.3: Top-*k* Accuracy for Test Set**

| k | Sample | Events | Sample and Events |
|---|--------|--------|-------------------|
| 1 | 0.571 | 0.585 | 0.668 |
| 2 | 0.614 | 0.622 | 0.707 |
| 3 | 0.635 | 0.641 | 0.725 |
| 4 | 0.65 | 0.655 | 0.74 |
| 5 | 0.662 | 0.666 | 0.749 |
| 6 | 0.672 | 0.674 | 0.757 |
| 7 | 0.682 | 0.682 | 0.764 |
| 8 | 0.689 | 0.688 | 0.77 |
| 9 | 0.697 | 0.693 | 0.776 |
| 10 | 0.704 | 0.699 | 0.781 |

**Table 4.4: Top-*k* Accuracy for Test Set for slide values that are a multiple of 2 seconds but not 4 seconds**

| k | Sample | Events | Sample and Events |
|---|--------|--------|-------------------|
| 1 | 0.452 | 0.58 | 0.667 |
| 2 | 0.509 | 0.621 | 0.705 |
| 3 | 0.537 | 0.641 | 0.727 |
| 4 | 0.556 | 0.654 | 0.739 |
| 5 | 0.573 | 0.663 | 0.749 |
| 6 | 0.587 | 0.672 | 0.756 |
| 7 | 0.6 | 0.68 | 0.762 |
| 8 | 0.611 | 0.685 | 0.766 |
| 9 | 0.621 | 0.691 | 0.772 |
| 10 | 0.629 | 0.695 | 0.776 |

## 4.5   Performance Analysis

As seen from the Figure 4.1 and Figure 4.3, the top-*k* accuracy for both training and test set is highest for the hash code type *sample and events*, followed by *events* and finally *sample*. Hence our event detection performs better than naive sampling technique. The main issue with sampling at a certain interval and taking respective hash code is that if there is a mismatch of selected time steps for hashing in the copied and original video, the expected results will be poor. However if you use event detection and choose event hashes as your hash values, the copied and original video will share similar hashes despite the mismatch of selected time steps for hashing since the concept of events are granular in nature. This claim is justified through our experimentation Figure 4.2 and Figure 4.4. When the slide value is not a multiple of $T_s$, we expect the *sample* to perform worse because in this case there is maximum mismatch between the frames. Figure 4.2 and Figure 4.4 show the result for train set and test set respectively with slide values that are a multiple of 2 seconds but not 4 seconds. As expected, the top-*k* accuracy of *sample* is much less than that obtained when all values of slide are considered, but there is not much difference in the accuracy of hash code type *sample and events* and *events*.

From Table 4.5 and Table 4.6, we observe that the average hash code length per seconds (AHL) obtained from *events* is lesser than that obtained from *sample* for most of the videos in train set and test set (except for those with video duration less than 15 seconds) and the AHL for *sample and events* is more or less same to that of *sample* for all the videos. Despite having smaller length hash codes in case of *events* and similar length hash code in case of *sample and events* compared to *sample*, *events* and *sample and events* yield better results in terms of Top-5 accuracy. The improved results of *sample and events* over *events* without much significant increase in length of hash code is due to the fact that as it is just adding extra hash codes when the event duration is very large. The increased AHL of *events* compared *sample* for small videos (duration less than 15 seconds) may be due to the fact that our model requires videos of certain minimum duration to detect events because of its hierarchical nature which calls for pooling of frames to get hash codes.

**Table 4.5: Average Hash code Length Per 5 seconds (AHL) and Top-5 Accuracy For Various Video Durations of Train Set**

| Video Duration (seconds) | Sample | | Events | | Sample and Events | |
|---|---|---|---|---|---|---|
| | AHL | Top-5 Acc. | AHL | Top-5 Acc. | AHL | Top-5 Acc. |
| <10 | 79.391 | 0.567 | 105.92 | 0.604 | 106.235 | 0.655 |
| 10-15 | 79.205 | 0.713 | 82.308 | 0.722 | 85.858 | 0.795 |
| 15-20 | 79.975 | 0.76 | 75.656 | 0.783 | 83.219 | 0.855 |
| 20-25 | 79.877 | 0.782 | 71.882 | 0.815 | 82.305 | 0.883 |
| 25-30 | 79.827 | 0.812 | 68.856 | 0.84 | 81.989 | 0.903 |
| 30-35 | 79.938 | 0.833 | 66.032 | 0.861 | 80.408 | 0.913 |
| 35-40 | 80.009 | 0.834 | 65.205 | 0.863 | 80.976 | 0.915 |
| 40-45 | 79.171 | 0.853 | 63.976 | 0.893 | 80.557 | 0.927 |
| 45-50 | 79.947 | 0.874 | 71.304 | 0.934 | 86.451 | 0.952 |
| >50 | 80.18 | 0.953 | 74.445 | 0.973 | 88.81 | 0.987 |

**Table 4.6: Average Hash code Length Per 5 seconds (AHL) and Top-5 Accuracy For Various Video Durations of Test Set**

| Video Duration (seconds) | Sample | | Events | | Sample and Events | |
|---|---|---|---|---|---|---|
| | AHL | Top-5 Acc. | AHL | Top-5 Acc. | AHL | Top-5 Acc. |
| <10 | 79.407 | 0.437 | 102.63 | 0.462 | 103.442 | 0.534 |
| 10-15 | 79.365 | 0.576 | 80.64 | 0.583 | 84.037 | 0.67 |
| 15-20 | 79.991 | 0.634 | 73.8 | 0.624 | 83.078 | 0.726 |
| 20-25 | 79.977 | 0.671 | 70.114 | 0.646 | 82.307 | 0.751 |
| 25-30 | 79.84 | 0.69 | 68.071 | 0.694 | 82.961 | 0.799 |
| 30-35 | 79.886 | 0.698 | 66.721 | 0.707 | 81.9 | 0.801 |
| 35-40 | 80.049 | 0.747 | 64.805 | 0.717 | 81.321 | 0.833 |
| 40-45 | 79.588 | 0.794 | 62.875 | 0.76 | 80.642 | 0.867 |
| 45-50 | 80.074 | 0.845 | 68.961 | 0.855 | 83.749 | 0.931 |
| >50 | 80.044 | 0.936 | 72.508 | 0.948 | 86.6 | 0.982 |

## 4.6 Qualitative Results

The main contribution of our work as verified by our experiments is how the concept of identifying events and hashing them lead to compact and efficient representation of hash codes for video segment copy detection with copies varying their length from very short till the length of full videos.This section explores quantitatively what an event is trying to represent based on experimental observations.

We have manually annotated where shot ends occurred; a shot is a sequence of frames that was (or appears to be) continuously captured from the same camera. Each plot in Figure 4.5-4.8 are the plots of hamming code distance between $d_{j,t}$ (the hamming distance between $b_{j,t+1}$ and $b_{j,t}$ as defined before) vs the frame number. The orange dots represents the manually annotated shot ends and the red dots are the end of events as detected by our model. As seen in Figure 4.5-4.8, for all the observed videos we have an end of an event nearby whenever there is a shot end. This is expected, as two events are expected to be independent of each other, and so there should be an event end whenever there is a shot end, as the next shot is independent of previous shot visually. However from Figure 4.5-4.8, we also observe that event ends (red dots) are occurring not only in the places where there are shots. This is due to the deeper understanding of the video by our model for identifying events. We now look into what these additional points try to convey by observing few such points manually.

In Figure 4.5.(a), around frame number 1151, we see an event being detected which is not a shot end. A few frames (frame 1121 to 1191) around this frame is plotted in Figure 4.9. As we can see, here the scene changes drastically around these frames as a man walks into the scene in these frames.

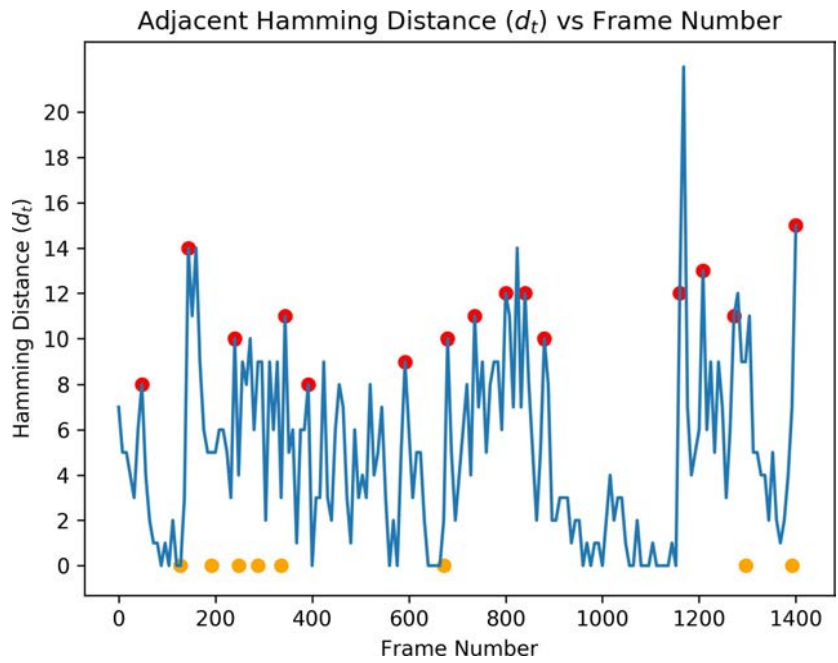

**(a)** Video 1

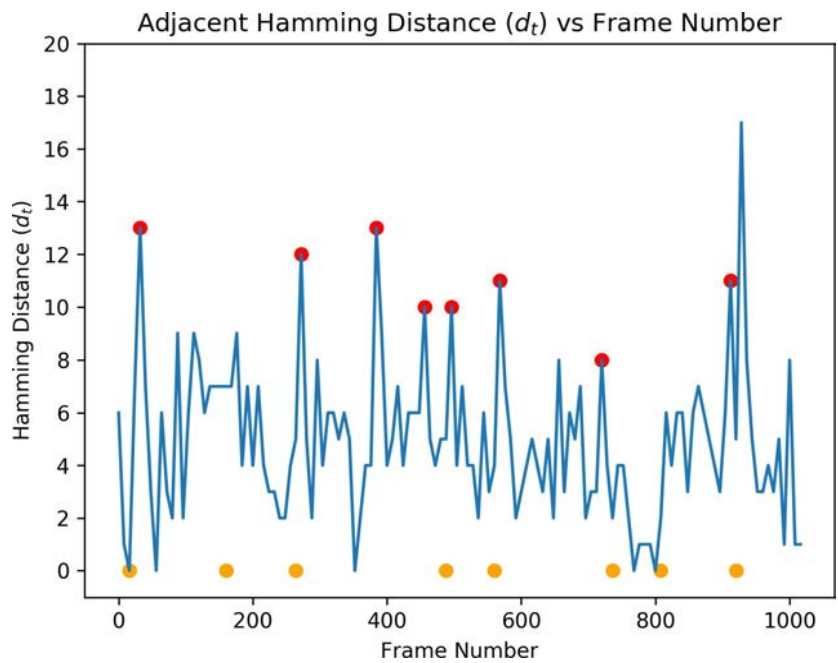

**(b)** Video 2

**Figure 4.5: Adjacent Hamming Distance ($d_t$) vs Frame Number for videos 1 and 2**

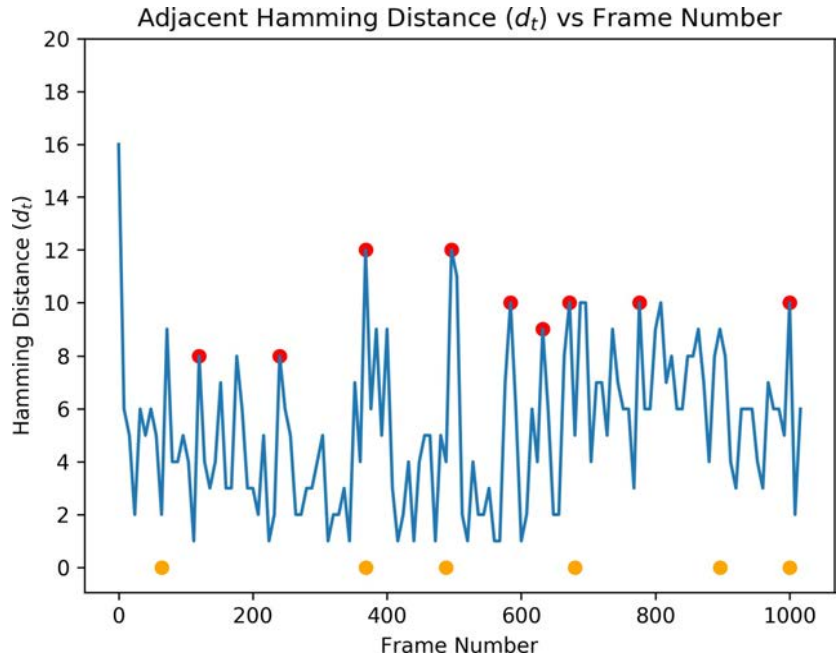

**(a)** Video 3

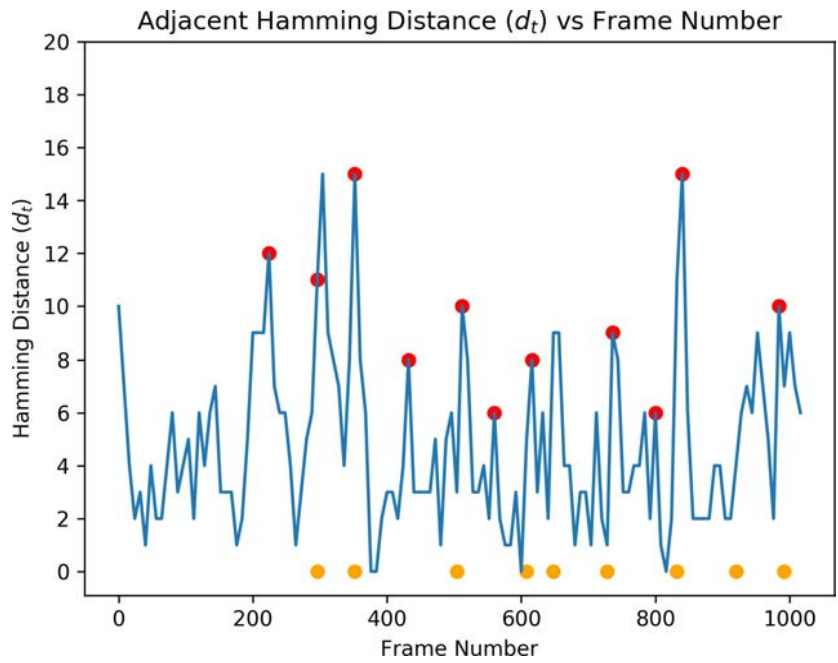

**(b)** Video 4

**Figure 4.6: Adjacent Hamming Distance ($d_t$) vs Frame Number for videos 3 and 4**

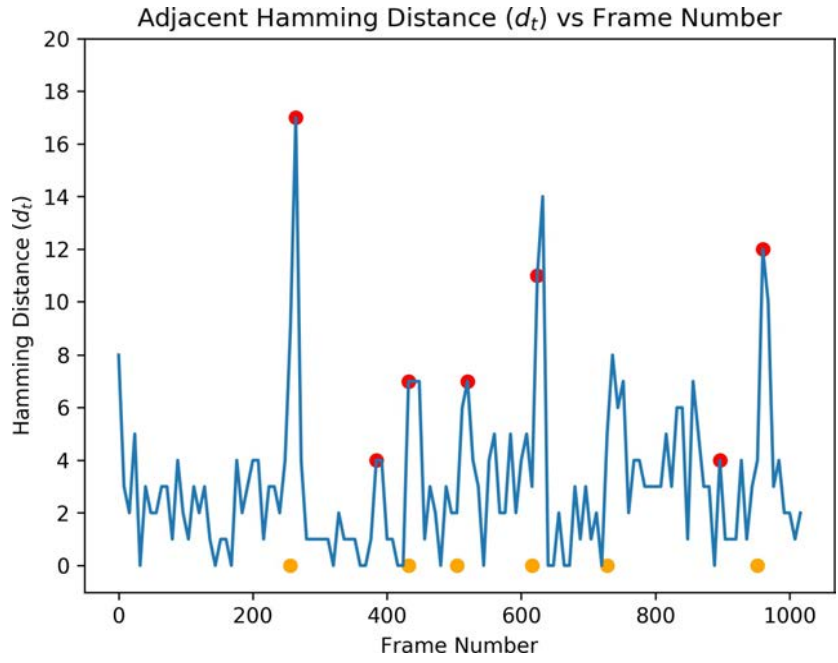

**(a)** Video 5

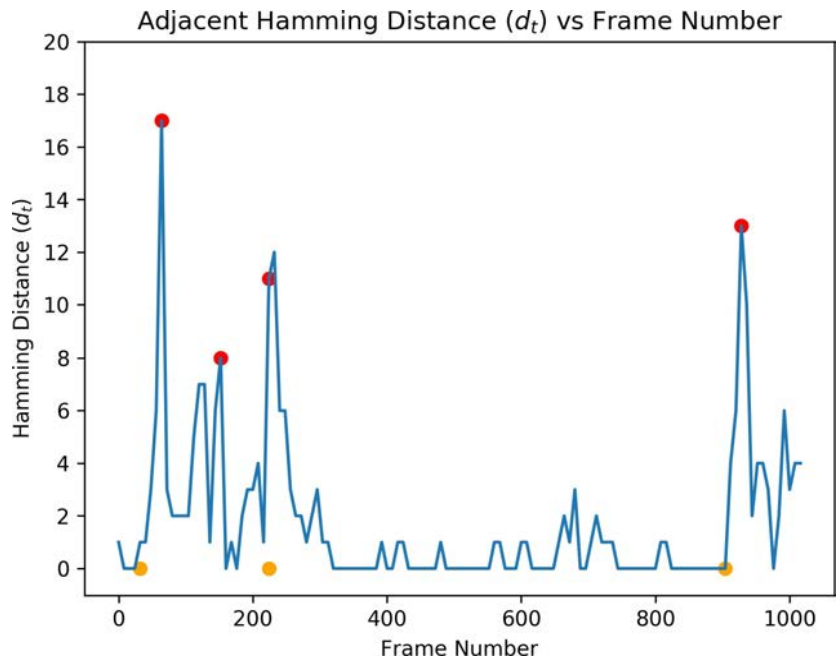

**(b)** Video 6

**Figure 4.7: Adjacent Hamming Distance ($d_t$) vs Frame Number for videos 5 and 6**

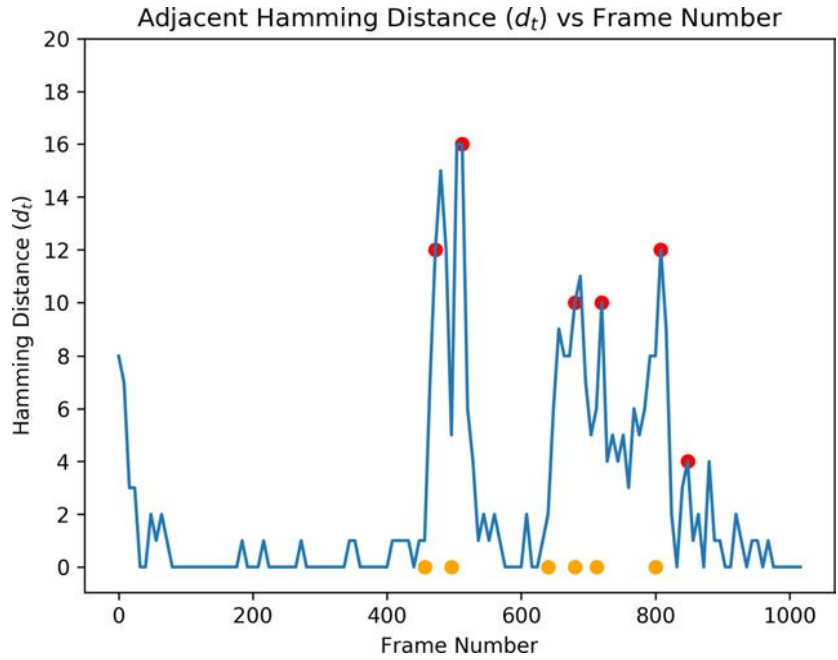

**(a)** Video 7

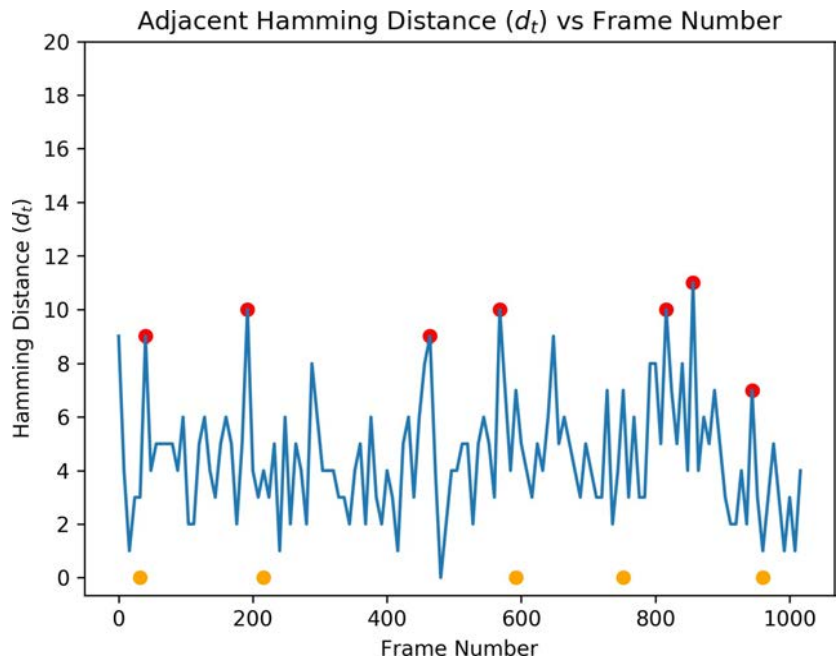

**(b)** Video 8

**Figure 4.8: Adjacent Hamming Distance ($d_t$) vs Frame Number for videos 7 and 8**

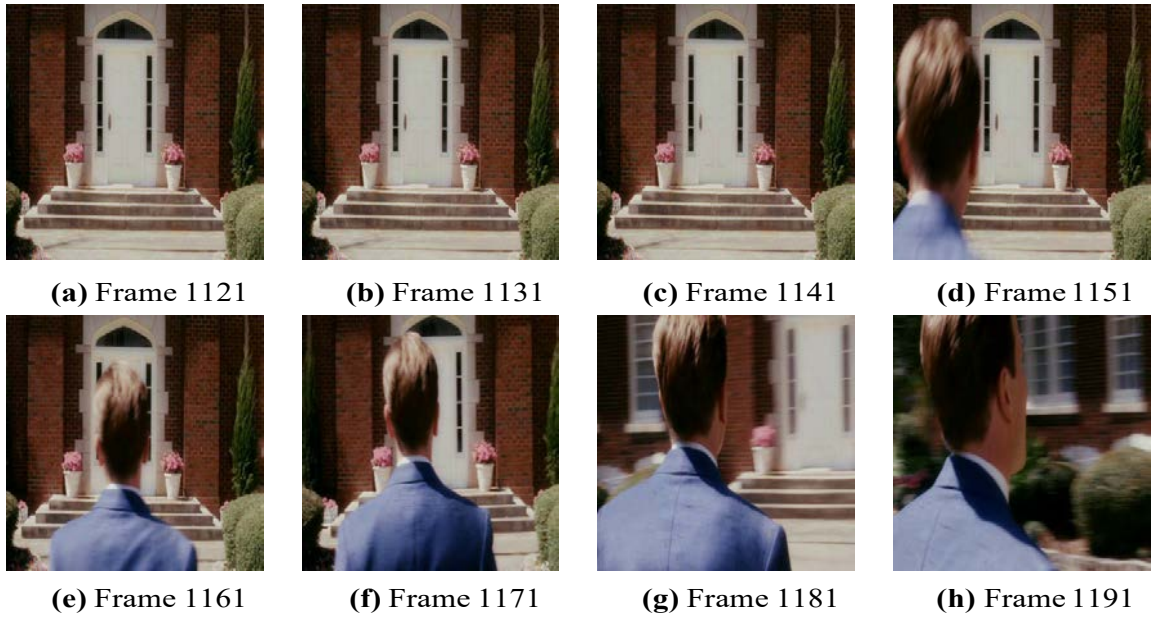

**(a)** Frame 1121    **(b)** Frame 1131    **(c)** Frame 1141    **(d)** Frame 1151

**(e)** Frame 1161    **(f)** Frame 1171    **(g)** Frame 1181    **(h)** Frame 1191

**Figure 4.9: Frames of Video 1**

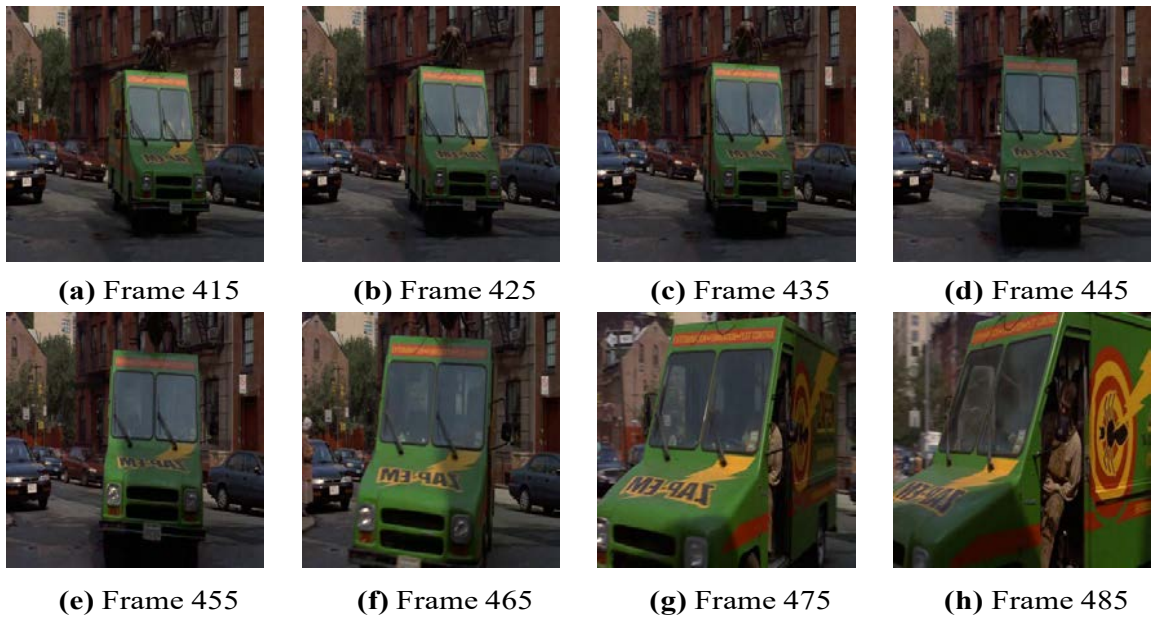

**(a)** Frame 415    **(b)** Frame 425    **(c)** Frame 435    **(d)** Frame 445

**(e)** Frame 455    **(f)** Frame 465    **(g)** Frame 475    **(h)** Frame 485

**Figure 4.10: Frames of Video 2**

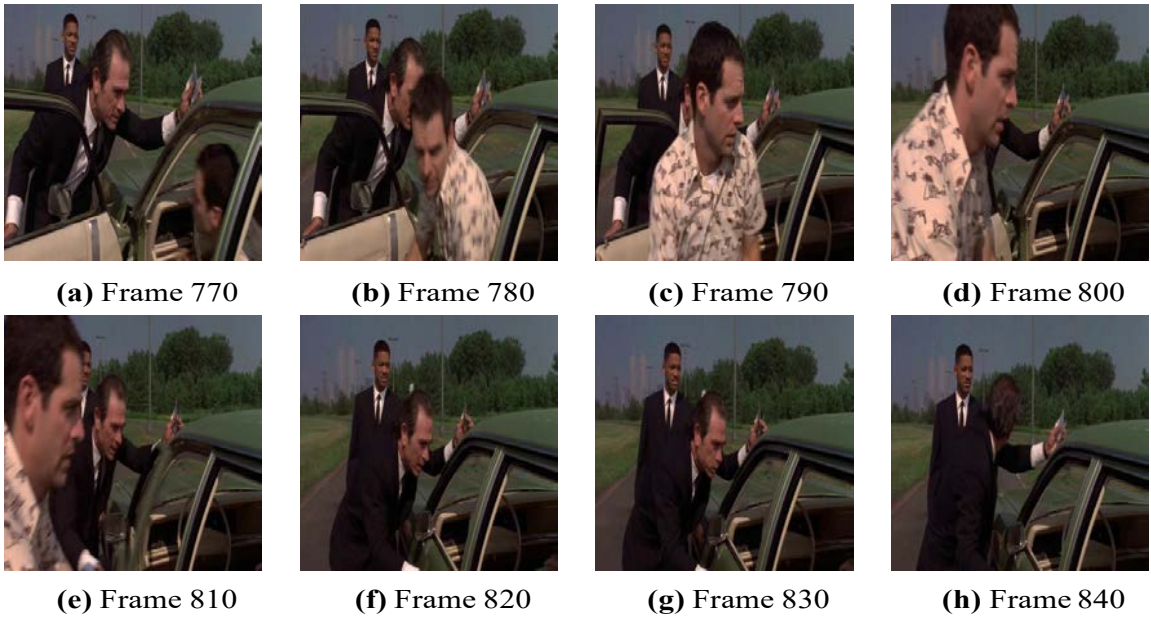

**(a)** Frame 770    **(b)** Frame 780    **(c)** Frame 790    **(d)** Frame 800

**(e)** Frame 810    **(f)** Frame 820    **(g)** Frame 830    **(h)** Frame 840

**Figure 4.11: Frames of Video 3**

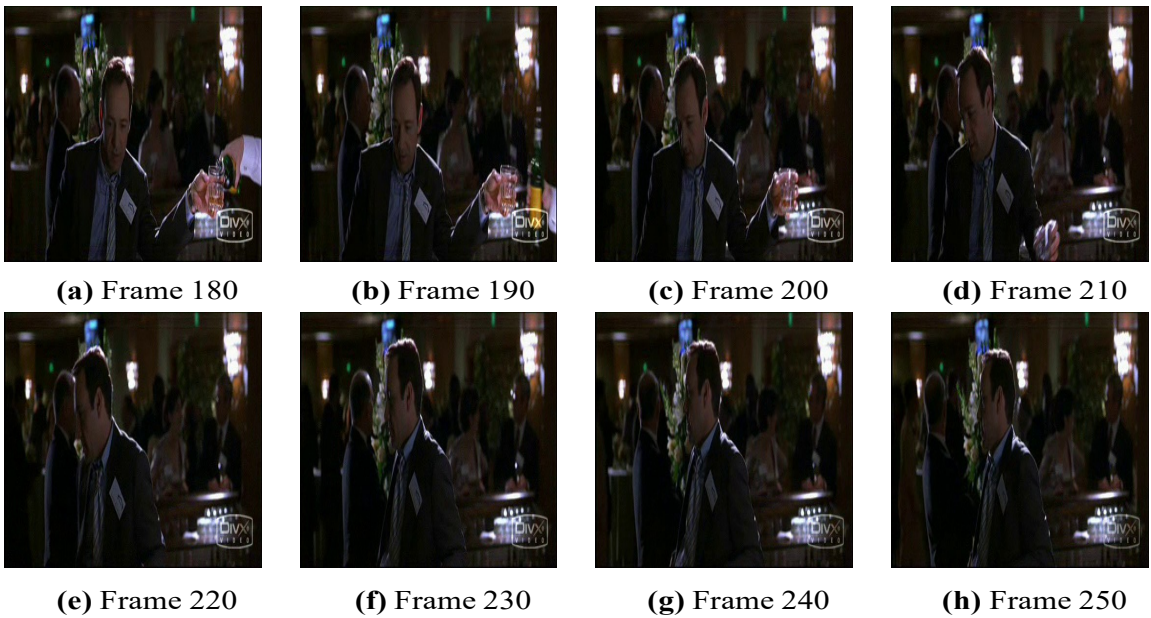

**(a)** Frame 180    **(b)** Frame 190    **(c)** Frame 200    **(d)** Frame 210

**(e)** Frame 220    **(f)** Frame 230    **(g)** Frame 240    **(h)** Frame 250

**Figure 4.12: Frames of Video 4**

Now consider Figure 4.5.(b). Around frame number 410, we see an event being detected which is not a shot end. A few frames (frame 415 to 485) around this frame is plotted in figure 4.10. From these frames, we observe that the object in this case is a van which is initially pretty far away from the camera. It then moves closer to the camera and changes the scene drastically around these frames, thereby leading to an event end that is detected here.

In Figure 4.6.(a), around frame number 780, we see an event being detected which is not a shot end. A few frames (frame 770 to 840) around this frame is plotted in Figure 4.11. From these frames, we observe that even though this is not a shot end, there are a lot of changes in these frames. These changes are happening even though they are continuously captured from the same camera, hence leading to an event end being detected here.

As can be seen in Figure 4.6.(b), around frame number 200, we see an event being detected which is not a shot end. A few frames (frame 180 to 250) around this frame is plotted in Figure 4.12. Here, the main object of the scene which is a man turning, leads to lot of information change between successive frames in this region. Therefore the information that these frames are packing cannot be contained in a hash code for one event, leading to an event end that is detected in this region.

## CONCLUSIONS

## 5.1   Conclusion

In this work, we have introduced the concept of events for compact and efficient representation of video hash codes for the purpose of video segment copy detection. We have proposed our novel deep neural network architecture, *Memory Constrained Hierarchical Batch-Normalized LSTM (MC BNLSTM) autoencoder*, for identifying and extracting hash codes for these events. To the best of our knowledge, MC BNLSTM autoencoder is the first method that learns the hash codes of a video by the reconstruction of the frame level content based features by an autoencoder architecture for the task of video copy detection. Experiments done on the difficult Hollywood2 dataset show that MC BNLSTM autoencoder can give very good retrieval accuracy even for small size video copy segments and is resilient to any frame level mismatches that may arise in a typical key frame based copy detection technique. However, we also find some shortcomings of our model. We have assumed that the video segment copies only contain cropped copy of the original video and not anything else. This assumption although is reasonable in case of information security applications involving sensitive videos, is not in general true for commercial and copyrighted video copies as these generally occur as partial copies, i.e. copied cropped video along with some other content. In the future, we will consider improving our hash code matching strategy to match the event hash codes created by our model to detect partial copies.

## 5.2   Scope for Future Research

As mentioned in the Sec.5.2, even though our proposed method works for video segment copies, it could be extended to a more general problem of partial copy detection by optimizing the hash code matching strategy. Another scope of extension of our work is to engineer more robust frame level feature such as CNN features which are used in [52], [55]-[57] to better deal with the frame level attacks done to the video such as frame in frame, geometrical distortions etc., which may occur in real world datasets for copy detection pertaining to commercial and copyrighted videos. Through this extension, we can experiment with real world datasets like [55] and hence produce a general framework for copy detection for information security, commercial licensing and copyrighting applications.